В.В. Кромер

**Двухпараметрическая модель длины слова "язык – жанр"**

Результаты параметризации текстов различных жанров на различных языках в соответствии с ранее предложенной автором математической моделью длины слова (измеряемой количеством слогов) на основе распределения Чебанова-Фукса с равномерным распределением параметра (Mathematical Model of Word Length on the Basis of the Cebanov-Fucks Distribution with Uniform Parameter Distribution, электронный архив Computer Science, http://arXiv.org/abs/cs.CL/0102026, 24 февраля 2001 г.), позволяют определить общие контуры двухпараметрической модели длины слова "язык – жанр". В ранее предложенной модели на основе распределения Чебанова-Фукса текст характеризуется 2 параметрами – средней длиной слова $\lambda_0$ и математическим ожиданием длины слова в начале ранжированного по убыванию частоты списка текстовых слов $\lambda_1$. Анализ найденных параметров $\lambda_0$ и $\lambda_1$ по текстам различных жанров на различных языках (в доступных данных наиболее последовательно представлены жанр писем и тексты на немецком языке) позволяет высказать предположение о существовании 2 инвариантных параметров, полностью характеризующих текст в рамках рассматриваемой модели, при этом первый параметр, равный $I = (\lambda_0 - 1)(\lambda_1 - \lambda_{1\min})$, является инвариантом языка, второй параметр $\alpha = \dfrac{\lambda_0 - \lambda_1}{\lambda_0 - \lambda_{1\min}}$ является инвариантом жанра. Значение $\lambda_{1\min}$ – нижнее значение предела $\lambda_0$. Обследованные данные позволяют сделать вывод, что это значение близко к 0,5. Параметры $I$ и $\alpha$ допускают лингвистическую интерпретацию. $I$ отражает степень синтетичности языка и меняется (на материале обследованных языков) от 0,08 (английский язык) до 0,84 (итальянский язык), составляя для немецкого языка 0,34. Параметр $\alpha$ характеризует степень завершенности синергетических процессов оптимизации кода (языка). При $\alpha = 0$ распределение Чебанова-Фукса с равномерным распределением параметра вырождается в распределение с фиксированным параметром. Вырожденным распределением описывается, в частности, структура слова в древних языках (например, латинском). Значению $\alpha = 1$ соответствует $\lambda_1 = \lambda_{1\,min}$ ($\lambda_{1\,min} = 0{,}5$ на основе наличных данных) и наиболее оптимальное распределение математического ожидания длины текстового слова в зависимости от его ранга, т.е. предельная оптимизация языка. Для реальных текстов $\alpha \in (0, 1)$, при этом для простых безыскусных жанров (писем) $\alpha \approx 0{,}6$, для изощренных (научные тексты, газетные тексты) $\alpha \approx 0{,}8$. Таким образом, в системе координат $I\alpha$ точки, отображающие тексты одного языка, располагаются вдоль вертикальных линий ($I$ = const.), а точки, отображающие одножанровые тек-

сты – вдоль горизонтальных линий ($\alpha$ = const.). На рисунке представлены точки $\alpha(I)$ для текстов в жанре писем (языки английский, французский, немецкий, шведский, испанский, итальянский) и разножанровых немецких текстов. Вертикальная и горизонтальная линии проведены в соответствии с усредненными значениями параметров $I$ и $\alpha$ для немецкого языка и жанра писем.

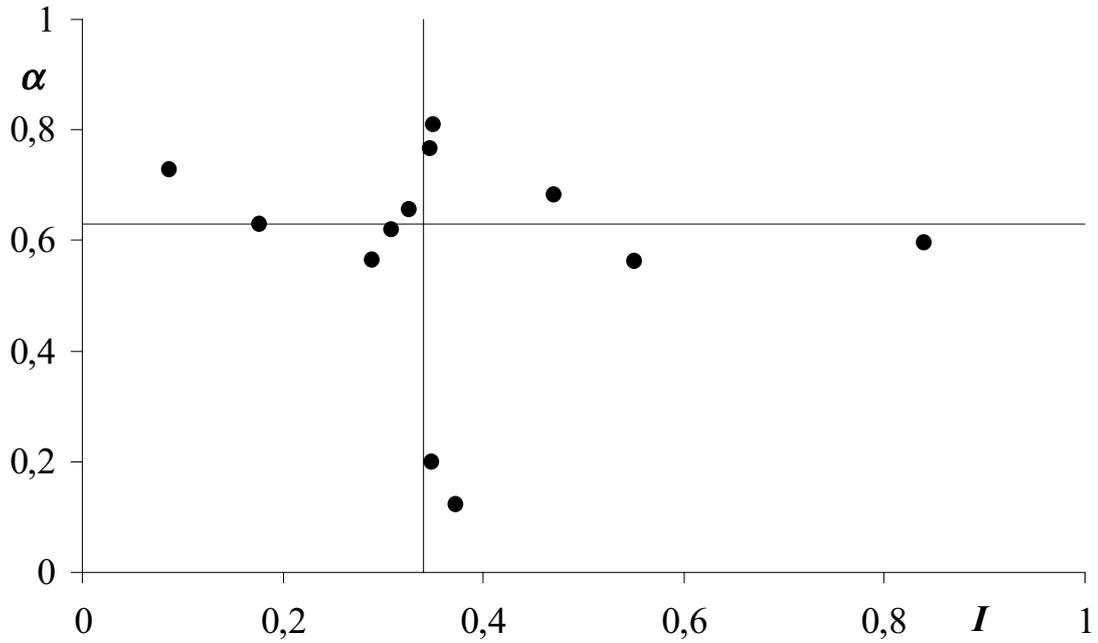